\documentclass[letterpaper, 10 pt, twocolumn, conference, numbers,sort&compress]{ieeeconf} 
\usepackage{rotating}

\IEEEoverridecommandlockouts                              

\let\labelindent\relax
\usepackage{graphicx}
\usepackage{booktabs}
\usepackage{cite}
\usepackage{amsmath}
\usepackage{algorithm}
\usepackage{algorithmic}
\usepackage{amssymb}
\usepackage{lipsum}
\usepackage{caption}
\usepackage{xcolor}
\usepackage{placeins}
\usepackage{booktabs} 
\usepackage{pifont}
\usepackage{multirow}
\usepackage{subfig}
\usepackage{hyperref}
\hypersetup{
    colorlinks = true,
    linkcolor = blue,
    urlcolor = blue,
}
\usepackage{enumitem}
\usepackage[T1]{fontenc}
\usepackage{CJKutf8}
\usepackage{url}
\usepackage{adjustbox}

\newcommand{\nop}[1]{}

\title{\LARGE \bf
Large-Scale Mixed-Traffic and Intersection Control using Multi-agent Reinforcement Learning 
}

\author{Songyang Liu$^{1}$, Muyang Fan$^{2}$, Weizi Li$^{3}$, Jing Du$^{1}$, Shuai Li$^{1}$
\thanks{$^{1}$Songyang Liu, Jing Du, and Shuai Li are with Department of Civil and Coastal Engineering at University of Florida, Gainesville, FL, USA {\tt\small liusongyang@ufl.edu; eric.du@essie.ufl.edu; shuai.li@ufl.edu}}%
\thanks{$^{2}$Muyang Fan is with Department of Computer Science at University of Memphis, Memphis, TN, USA {\tt\small mfan1@memphis.edu}}%
\thanks{$^{3}$Weizi Li is with Min H. Kao Department of Electrical Engineering and Computer Science at University of Tennessee, Knoxville, TN, USA {\tt\small weizili@utk.edu}}%
}
\begin{document}
\begin{CJK}{UTF8}{gbsn}

\maketitle
\thispagestyle{empty}
\pagestyle{empty}

\begin{abstract}
Traffic congestion remains a significant challenge in modern urban networks. Autonomous driving technologies have emerged as a potential solution.
Among traffic control methods, reinforcement learning has shown superior performance over traffic signals in various scenarios. 
However, prior research has largely focused on small-scale networks or isolated intersections, leaving large-scale mixed traffic control largely unexplored. This study presents the first attempt to use decentralized multi-agent reinforcement learning for large-scale mixed traffic control in which some intersections are managed by traffic signals and others by robot vehicles. Evaluating a real-world network in Colorado Springs, CO, USA with 14 intersections, we measure traffic efficiency via average waiting time of vehicles at intersections and the number of vehicles reaching their destinations within a time window (i.e., throughput). 
At 80\% RV penetration rate, our method reduces waiting time from 6.17~$s$ to 5.09~$s$ and increases throughput from 454 vehicles per 500 seconds to 493 vehicles per 500 seconds, outperforming the baseline of fully signalized intersections. These findings suggest that integrating reinforcement learning-based control large-scale traffic can improve overall efficiency and may inform future urban planning strategies. 
\end{abstract}


\section{Introduction}


Traffic congestion remains a major issue in modern urban systems. In 2024, it caused drivers to lose an average of 43 hours in traffic, costing \$771 in lost time and productivity. New York City, Chicago (both 102 hours), and Los Angeles (88 hours) were the most congested U.S. cities, costing drivers \$1,826 and \$1,575, respectively~\cite{INRIX_US}. Similarly, London drivers faced 101 hours of congestion in 2024, costing the city \$4.88 billion, or \$1,193 per driver~\cite{INRIX_UK}. These figures highlight congestion’s prevalence and high costs worldwide.

Autonomous driving technologies have been explored to alleviate congestion. Rapid progress has led to more vehicles being equipped with autonomous features and tested in real-world conditions. 
Nevertheless, the shift toward full transportation autonomy is expected to experience a prolonged period in which human-operated vehicles (HVs) and robot vehicles (RVs) cooperate, hence mixed traffic. 

Traffic control methods have also seen significant progress in recent years. 
In particular, Reinforcement learning (RL)-based vehicle control methods have been shown to outperform traffic signals at complex intersections~\cite{yan2021reinforcement, wang2024learning,Villarreal2024Eco,
Islam2024Heterogeneous}. Studies have demonstrated the feasibility of using image data rather than detailed traffic measurements for RL-based control~\cite{villarreal2024mixed}, developed RL frameworks incorporating real-world driving profiles to enhance safety, stability, and efficiency in mixed traffic scenarios~\cite{poudel2024carl, poudel2024endurl}, and provided evidence that RL-based techniques can reduce collisions and improve overall urban traffic flow~\cite{peng2021connected, shi2022control}.
Despite these advancements, existing research has been limited to small-scale settings, with little exploration of large-scale mixed-traffic control where traffic signals and RVs operate together. Recent multi-agent frameworks~\cite{spatharis2024multiagent} and control methods for mixed traffic~\cite{shi2022control} suggest that integrating these approaches may be feasible, yet a solution for large-scale traffic remains unexplored. 


We present the first attempt of applying decentralized multi-agent reinforcement learning (MARL) for large-scale mixed-traffic and intersection control, in which traffic signals govern some intersections while RL-controlled RVs regulate the others. Unlike previous studies that focus exclusively on solely signalized or unsignalized control, we investigate how RVs and traffic signals can coexist to optimize urban mobility. 
Using our MARL-based control method, RVs adjust their behavior dynamically based on local traffic conditions, offering a flexible alternative to static traffic signals. 


We conduct experiments on a large-scale network of 14 intersections using the Simulation of Urban MObility (SUMO)\cite{lopez2018microscopic}. 
To evaluate our approach, we test five configurations of 14 intersections. 
Each configuration divides the 14 intersections into a set of unsignalized intersections controlled by RL-driven RVs and a set of signalized intersections that are regulated by traffic lights.
We compare this mixed-control approach with a baseline where all intersections rely solely on traffic signals. We evaluate traffic performance using two metrics: 1) average waiting time $\overline{W}$, which measures traffic efficiency by averaging the total time vehicles remain motionless near an intersection~\cite{zhang2020using, greguric2020application, wang2024learning}, and 2) vehicle throughput~$Q$, which represents the number of vehicles crossing an intersection or reaching their destinations over the second half of an episode (500 seconds).
The most significant improvement over the baseline is observed when the RV penetration rate reaches 80\%: $\overline{W}$ reduces from 6.17~$s$ to 5.09~$s$ (a 20\% deduction) and~$Q$ increases from 454 vehicles per 500 seconds ($v/500 s$) to 493~$v/500 s$ (a 9\% improvement). 

By bridging the gap between fully signalized and autonomous traffic control, this study introduces a new paradigm for urban traffic management. As trends in autonomous transportation continue, our findings may inform urban planning strategies by showing that mixed-traffic control can improve efficiency while reducing reliance on static signals. This work is an initial step toward understanding the feasibility and effectiveness of a large-scale transition to signal-free intersections in real-world networks.
The Code of our work is available at~\href{https://github.com/cgchrfchscyrh/MixedTrafficControl\_IROS}{https://github.com/cgchrfchscyrh/MixedTrafficControl\_IROS}.
\section{Related Work}
Managing urban intersections has become increasingly challenging as autonomous and human-driven vehicles continue to share the roads.  Existing traffic control techniques, such as fixed-time and actuated signaling have been extensively implemented to regulate traffic under stable conditions~\cite{NACTO, Greenlight}. 
However, these methods often struggle to cope with the dynamic and complex nature of modern traffic environments~\cite{gholamhosseinian2022comprehensive}. Meanwhile, optimization-based methods, including integer linear programming and rule-based strategies, can improve efficiency but encounter  scalability issues when applied to expansive networks~\cite{qadri2020state}. 
In contrast, learning-based control methods can dynamically adapt to changing traffic conditions, optimizing vehicle movement in real time. Following this approach, we develop an RL-based vehicle control strategy to enhance traffic efficiency.

Various studies have focused on controlling traffic at unsignalized intersections, especially for connected and autonomous vehicles (CAVs). For example, an early study proposed a multi-agent control system in which CAVs secure space-time reservations on a first-come, first-served basis~\cite{dresner2008multiagent}. Building on this idea, another study applied the FCFS scheduling strategy to CAV platoons~\cite{jin2013platoon}, and a controllable gap method was introduced that dynamically adjusts the time intervals between vehicles—accounting for their speeds and potential conflicts to reduce collision risks~\cite{chen2022improved}. Additional research has explored decentralized approaches, such as consensus-based trajectory management~\cite{mirheli2019consensus} and energy optimization techniques~\cite{malikopoulos2018decentralized}, although these methods typically assume an environment composed entirely of autonomous vehicles. 
In addition, these approaches are based on the assumption that all vehicles are connected, which does not hold in mixed-traffic conditions where HVs lack connectivity. 
Hence, our method assumes only RVs can exchange traffic information within a range of intersections, allowing for a more realistic and adaptable control strategy.


Recent studies suggest that RL can enable RVs to influence HVs and optimize traffic efficiency~\cite{adu2014application, wu2021flow, yan2021reinforcement, you2019advanced, wang2024learning, villarreal2024mixed,Villarreal2023Chat,niroumand2023white, ferrarotti2024autonomous}, indicating that RV-based control can outperform traffic signals at unsignalized intersections~\cite{yan2021reinforcement, wang2024learning}. Several studies demonstrate that RL-driven control---whether through decision-making models~\cite{xu2022decision} or multi-task strategies~\cite{kai2020multi}---can outperform traffic signals in unsignalized settings. 
Moreover, studies focusing on efficient RL methods~\cite{wang2023efficient} and curriculum-based approaches~\cite{khaitan2022state}, along with enhanced strategies for urban traffic scenarios~\cite{yin2023efficient}, have shown promising improvements in convergence and performance. Specifically, leveraging expert priors and structured policy exploration improves sample efficiency, while techniques such as value distribution estimation and skill-based learning contribute to the stability of RL-based control. 

Decentralized multi-agent RL has been shown to substantially reduce waiting times at complex unsignalized intersections, particularly when there is a high presence of RVs~\cite{wang2023large, wang2024learning}. One study proposed a decentralized RL framework for large-scale traffic signal management leveraging localized optimization and indirect collaboration through shared state information~\cite{ma2024efficient}, while another extended RL coordination to city-wide mixed traffic networks by incorporating dynamic vehicle routing and privacy-preserving crowdsourcing to balance RV penetration and mitigate localized congestion~\cite{wang2023large}. 
Unlike earlier approaches that depend on either traffic lights or full autonomous coordination, our method explores how both systems can operate together to optimize urban mobility.


\section{Methodology}
We first explain our mixed control method at intersections. Second, we introduce our vehicle behavior modeling method, followed by the implementation details of our method. 
Lastly, we introduce the evaluation metrics of our approach.

\subsection{RL-based Control for Mixed Traffic at Intersections}
We address the challenge of coordinating mixed traffic at unsignalized intersections by formulating the task as a Partially Observable Markov Decision Process (POMDP): 
\[
(S, A, T, R, O, Z, \mu_0, \gamma),
\]
where 
\begin{itemize}
    \item $S$ is the set of all possible states;
    \item $A$ is the set of actions available to the vehicles;
    \item $T(s' \mid s, a)$ denotes the probability of transitioning from state $s$ to state $s'$ when action $a$ is taken;
    \item $R(s, a)$ is the reward function providing feedback for a given state-action pair;
    \item $O$ represents the observation space corresponding to the partial view of the environment available to each RV;
    \item $Z(o \mid s)$ gives the likelihood of obtaining observation $o$ when the true state is $s$;
    \item $\mu_0$ is the distribution of the initial state; and
    \item $\gamma$ is the discount factor applied to future rewards.
\end{itemize}

At each discrete time step $t$, an RV selects an action $a_t \in A$ according to a policy $\pi_\theta(a_t \mid s_t)$. Upon taking this action, the environment updates to a new state $s_{t+1}$ and the RV receives an immediate reward $r_t$. In our multi-agent setting, all RVs share the same policy while making decisions independently. 
We seek to maximize the overall discounted reward $G_t$, which sums the immediate rewards from time~$t$ to~$T$, each multiplied by a discount factor~$\gamma^{\,j-t}$. 


Each RV is faced with a binary decision Stop or Go. The Stop command prevents the vehicle from entering the intersection, whereas the Go command permits it to continue moving. At time $t$, the observation vector $o_t$ that a RV receives is structured as

\begin{equation}
    o_t = \bigoplus_{d \in D} \langle q_d, \tau_d \rangle \bigoplus_{d \in D} \sigma_d,
\end{equation}

\noindent where~$D$ represents the set of all traffic directions, while~$q_d$ denotes the number of vehicles queued in a specific direction~$d$. The variable~$\tau_d$ corresponds to the average waiting time of vehicles arriving from direction~$d$, and~$\sigma_d$ indicates whether vehicles from direction~$d$ are currently occupying the intersection.

To promote efficient traffic flow and mitigate potential conflicts, we design a reward function that blends local performance with a penalty for conflicts. Conflicts between robot vehicles occur when multiple RVs attempt to enter or pass through an intersection at the same time in ways that lead to overlapping or intersecting paths, such as left-turn and straight-going movements crossing each other.
Specifically, when vehicles approaching from conflicting directions converge at the intersection, priority is assigned based on a weighted measure of the queue length and waiting time. The reward received at time $t$ is defined as

\begin{equation}
    r_t = \beta \cdot r^{\text{local}}_t + r^{\text{penalty}}_t,
\end{equation}

\noindent where $\beta$ is a scaling constant. At each time step, the local reward is determined by the vehicle's action. 
If the vehicle executes Go, it receives a reward equal to~$\tau_d$, the average waiting time for the relevant direction. If it instead chooses Stop, it receives a penalty of~$-\tau_d$. In addition, if a conflict occurs at the intersection, the vehicle incurs a penalty of~$-1$; otherwise, there is no penalty.





\subsection{Vehicle Behavior Modeling}
HVs are simulated using the Intelligent Driver Model (IDM)~\cite{treiber2000congested}, which calculates acceleration based on the current traffic environment. In contrast, RVs employ a hybrid strategy: they follow IDM rules when they are more than 30 meters away from the intersection, and switch to a learned control policy once they are within the intersection zone.

When the learned policy directs an RV to Go, the vehicle accelerates at its maximum allowable rate. However, if the policy mandates Stop, the vehicle decelerates in a manner that depends on its current speed $v$ and the distance remaining to the intersection, denoted as $d_{\text{int}}$. 
The deceleration is calculated as $-v^2/2\,d_{\text{int}}$, where~$v$ is the current speed of vehicle, and~$d_{\text{int}}$ is the distance to the intersection. 




\subsection{Mixed Control of Intersections}
We implement two distinct approaches to intersection control: one that uses RL-controlled RV coordination and the other that relies on traffic signals. At intersections managed by RVs, the traffic lights are disabled, enabling RVs to regulate traffic. 
Our study investigates how these two methods can coexist within a large-scale traffic network to optimize urban mobility. 

We evaluate five configurations, denoted as~$xU+yS$, where~$x$ is the number of intersections controlled by RVs and~$y$ is the number regulated by traffic signals,~$x \in \{0,2,4,6,8,10\}$,~$y \in \{14,12,10,8,6,4\}$, totaling 14 intersections. For example, in the~$2U+12S$ setup, two intersections are managed by RVs and 12 by traffic signals. 
For comparison, the baseline configuration~$0U+14S$ represents all 14 intersections being managed by traffic signals. This design enables the study of the impact of shifting control from traffic signals to RVs on large-scale traffic.

\subsection{Evaluation Metrics}
We assess traffic performance using the following metrics.

\subsubsection{Average Waiting Time ($\overline{W}$)}
This metric has been adopted frequently to evaluate traffic efficiency over a road network~\cite{zhang2020using, greguric2020application, wang2024learning}. For each vehicle, the waiting time is defined as the total duration it remains motionless within the control zone. To compute this metric, we first determine the average waiting time for a specific moving direction by averaging the waiting times of all vehicles traveling in that direction; similarly, for an intersection, we average the waiting times of all vehicles present.~$\overline{W}$ is computed by adding up every vehicle's waiting time and then dividing that sum by the number of vehicles. This metric serves as an indicator of traffic efficiency, with lower values reflecting improved flow and reduced congestion.



\subsubsection{Vehicle Throughput ($Q$)} Vehicle throughput is defined differently at the individual intersection level and the network level. For a single intersection, $Q$ represents the number of vehicles that successfully pass that intersection. 
For an entire network, $Q$ refers to the total number of vehicles that reach their designated destinations. This metric provides insight into both localized traffic efficiency at intersections and overall traffic performance network-wide.
The higher the throughput, the smoother the traffic flow and better the system optimization.

\section{Experiments and Results}
We first explain our experiment set-up, then the traffic performance at individual intersections and the entire network.

\subsection{Experiment Set-up}
\begin{figure}
    \centering
    \includegraphics[width=\linewidth]{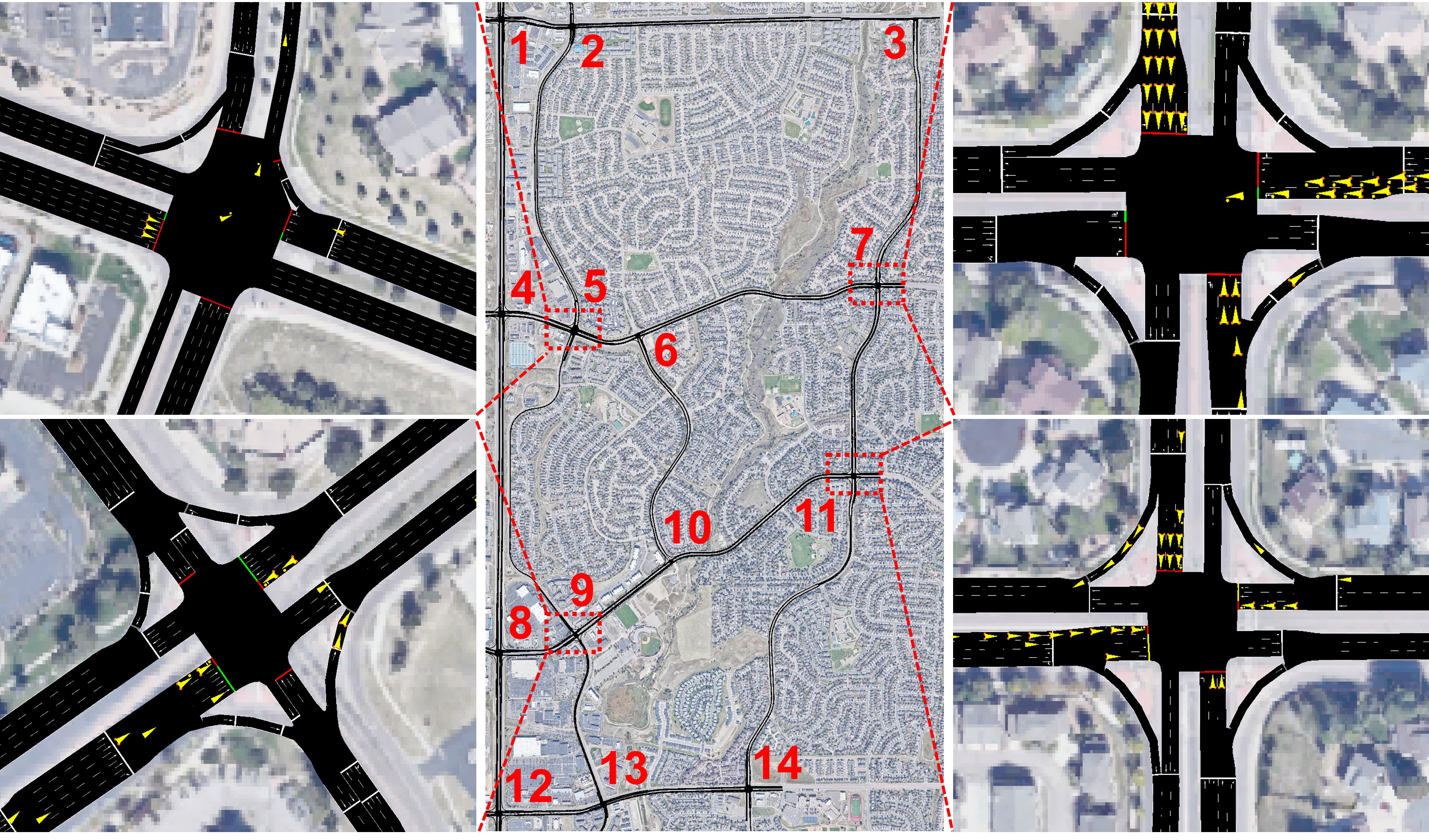}
    \caption{\small{We study a large-scale network that consists of 14 intersections in Colorado Springs, CO, USA. Four intersections (i.e., 5, 7, 9, and 11) are shown in close-up views as examples.}}
    \label{fig:Fig_1}
    \vspace{-16pt}
\end{figure}

Our simulation network shown in Fig.~\ref{fig:Fig_1} is constructed using intersection data from the city of Colorado Springs, CO, USA. The network features 12 peripheral entrances and exits (where origin-destination pairs are defined) and 14 intersections. 
We experiment with various configurations by designating a subset of intersections (i.e., 2, 4, 6, 8, or 10) to be managed by RVs, while the remaining intersections operate under traffic signals. 

For training our RL policy, we employ the Rainbow DQN algorithm~\cite{hessel2018rainbow} over 1,000 iterations. We vary the RV penetration rates from 40\% to 80\% in 10\% increments, with each setting requiring between 10 to 30 hours of training using 16 CPU cores from an AMD EPYC 7742 Processor and an NVIDIA A100 graphics card. 
The prioritized replay buffer parameter~$\alpha$ is set to 0.5, with a replay buffer capacity of 50,000. The algorithm utilizes 51 atoms and a neural network with three hidden layers of 512 neurons each. A discount factor of 0.99 is applied to prioritize long-term rewards, while a mini-batch size of 32 is used for training updates. The learning rate is set at 0.0005 to control the step size during optimization. Additionally, the control zone radius is defined as 30 meters, determining the spatial range for decision-making in the environment.
After training, each RL policy is evaluated 100 times, and the results are reported using the values over these 100 evaluations, in which each evaluation runs for a 1000-second simulation period.



\subsection{Overall Results}

\begin{table*}[ht!]
\vspace{6pt}
\centering
\setlength{\tabcolsep}{5.75pt} 
\renewcommand{\arraystretch}{1.35} 
\begin{center}
\begin{adjustbox}{max width=\textwidth}{
    \begin{tabular}{|c|cc|cc|cc|cc|cc|cc|cc|}
        \hline
        & \multicolumn{12}{c|}{Intersection control configuration}\\
        \cline{2-13}
        Intersection
        & \multicolumn{2}{c|}{$0U+14S$ (baseline)}
        & \multicolumn{2}{c|}{$2U+12S$}
        & \multicolumn{2}{c|}{$4U+10S$}
        & \multicolumn{2}{c|}{$6U+8S$}
        & \multicolumn{2}{c|}{$8U+6S$}
        & \multicolumn{2}{c|}{$10U+4S$} \\ 
        \cline{2-13}
        & $\overline{W}~(s)$ & $Q~(v/500s)$
        & $\overline{W}~(s)$ & $Q~(v/500s)$
        & $\overline{W}~(s)$ & $Q~(v/500s)$
        & $\overline{W}~(s)$ & $Q~(v/500s)$
        & $\overline{W}~(s)$ & $Q~(v/500s)$
        & $\overline{W}~(s)$ & $Q~(v/500s)$\\
        \hline
1
        & 5.27 & 391
        & 0.53 & 349
        & 0.67 & 346
        & 0.67 & 346
        & 0.6 & 351
        & 0.63 & 355 \\

        \hline
2
        & 3.07 & 194
        & 4.92 & 190
        & 4.97 & 191
        & 17.47 & 170
        & 6.9 & 192
        & 16.4 & 178 \\

        \hline
3
        & 7.26 & 464
        & 7.12 & 362
        & 6.24 & 373
        & 8.63 & 332
        & 7.41 & 367
        & 7.15 & 371 \\

        \hline
4
        & 0.95 & 56
        & 2.49 & 64
        & 2.5 & 64
        & 6.2 & 56
        & 0.81 & 65
        & 6.42 & 61 \\

        \hline
5
        & 9.74 & 515
        & 9.65 & 504
        & 11.55 & 407
        & 11.62 & 386
        & 11.47 & 396
        & 11.82 & 382 \\

        \hline
6
        & 5.10 & 88
        & 3.03 & 87
        & 4.41 & 86
        & 4.37 & 89
        & 0.11 & 102
        & \textbf{0.1} & 108 \\

        \hline
7
        & 12.89 & 488
        & 13.22 & 490
        & 20.47 & 396
        & 21.92 & 367
        & 19.98 & 388
        & 18.31 & 393 \\

        \hline
8
        & 7.62& 198
        & 6.76 & 182
        & 7.24 & 184
        & 7.3 & 184
        & 4.2 & 177
        & 4.97 & 177 \\

        \hline
9
        & 1.91 & 190
        & 2.04 & 188
        & 2.0 & 191
        & 1.95 & 191
        & 2.02 & 191
        & 1.96 & 192 \\

        \hline
10
        & 0.28 & 39
        & 0.35 & 44
        & 0.37 & 43
        & 0.36 & 41
        & 0.4 & 43
        & 0.01 & 44 \\

        \hline
11
        & 17.83 & 144
        & 15.91 & 143
        & 16.81 & 138
        & 16.36 & 143
        & 16.45 & 145
        & 16.6 & 141 \\

        \hline
12
        & 0.79 & 112
        & 0.78 & 114
        & 0.81 & 117
        & 0.8 & 118
        & 1.1 & 121
        & 0.15 & 128 \\

        \hline
13
        & 4.19 & 221
        & 4.94 & 230
        & 5.03 & 233
        & 4.94 & 235
        & 5.2 & 235
        & 5.49 & 236 \\

        \hline
14
        & 9.49 & 354
        & 8.24 & 358
        & 8.47 & 359
        & 8.2 & 358
        & 8.21 & 363
        & 8.4 & 362 \\

        \hline
        Network  
        & 6.17 & 454  
        & \textbf{5.86} & 470 
        & 6.63 & 459 
        & 7.92 & 448
        & 6.11 & 471 
        & 7.02 & \textbf{473}\\

        \hline
    \end{tabular}
}
\end{adjustbox} 
\end{center}
\vspace{-8pt}
\caption{\small{Average waiting time $\overline{W}$ and vehicle throughput $Q$ for 14 individual intersections and entire network under six different network control configurations when RV rate = 50\%. Rows 1-14 show individual intersection data, while the last row represents the entire network’s data. For a single intersection, $Q$ represents the number of vehicles that successfully leave this intersection. For the entire network, $Q$ refers to the total number of vehicles that reach their designated destinations. Increasing the number of unsignalized intersections leads to variable performance. The configurations~$2U+12S$ and~$8U+6S$ achieve lower network~$\overline{W}$ and higher network~$Q$, while~$4U+10S$,~$6U+8S$ and~$10U+4S$ get higher~$\overline{W}$ and higher~$Q$ compared to the baseline~$0U+14S$. For the network, bold
values indicate the highest improvement for the two metrics, occurring at~$2U+12S$ and~$10U+4S$ configuration. Our method impacts individual intersections differently: some achieve lower~$\overline{W}$ and others experience higher~$\overline{W}$. For intersection 6, bold value indicates the highest improvement with an 98.04\% reduction in~$\overline{W}$. $2U+12S$ achieves the lowest~$\overline{W}$ of 5.86~$s$ across the entire network and has the lowest $\overline{W}$ at six of the 14 intersections, making it the best configuration for minimizing delays. For~$Q$,~$10U+4S$ achieves the highest network-wide value of 473~$v/500s$ and has the highest $Q$ at six intersections, while the baseline ranks second with four intersections. These results indicate that~$10U+4S$ is the best configuration for maximizing throughput.}}
\label{tab:50results}
\vspace{-10pt}
\end{table*}

\begin{table*}[ht!]
\vspace{6pt}
\centering
\setlength{\tabcolsep}{5.75pt} 
\renewcommand{\arraystretch}{1.35} 
\begin{center}

\begin{adjustbox}{max width=\textwidth}{
    \begin{tabular}{|c|cc|cc|cc|cc|cc|cc|cc|}
        \hline
        & \multicolumn{12}{c|}{Intersection control configuration}\\
        \cline{2-13}
        Intersection
        & \multicolumn{2}{c|}{$0U+14S$ (baseline)}
        & \multicolumn{2}{c|}{$2U+12S$}
        & \multicolumn{2}{c|}{$4U+10S$}
        & \multicolumn{2}{c|}{$6U+8S$}
        & \multicolumn{2}{c|}{$8U+6S$}
        & \multicolumn{2}{c|}{$10U+4S$} \\ 
        \cline{2-13}
        & $\overline{W}~(s)$ & $Q~(v/500s)$
        & $\overline{W}~(s)$ & $Q~(v/500s)$
        & $\overline{W}~(s)$ & $Q~(v/500s)$
        & $\overline{W}~(s)$ & $Q~(v/500s)$
        & $\overline{W}~(s)$ & $Q~(v/500s)$
        & $\overline{W}~(s)$ & $Q~(v/500s)$\\
        \hline
1
        & 5.27 & 391
        & 0.46 & 359
        & 0.74 & 365
        & 0.77 & 364
        & 0.75 & 365
        & 0.75 & 364 \\

        \hline
2
        & 3.07 & 194
        & 4.83 & 190
        & 4.94 & 191
        & 8.46 & 190
        & 0.23 & 212
        & 0.69 & 206 \\

        \hline
3
        & 7.26 & 464
        & 3.89 & 430
        & 5.26 & 419
        & 6.11 & 398
        & 5.57 & 408
        & 6.9 & 378\\

        \hline
4
        & 0.95 & 56
        & 2.55 & 64
        & 2.46 & 63
        & 0.67 & 64
        & 0.06 & 65
        & 1.32 & 67 \\

        \hline
5
        & 9.74 & 515
        & 10.64 & 493
        & 8.93 & 435
        & 8.4 & 419
        & 8.83 & 437
        & 10.98 & 405 \\

        \hline
6
        & 5.10 & 88
        & 4.5 & 86
        & 4.3 & 87
        & 4.29 & 87
        & \textbf{0.08} & 99
        & 0.11 & 106 \\

        \hline
7
        & 12.89 & 488
        & 13.87 & 490
        & 16.08 & 431
        & 16.81 & 419
        & 15.76 & 432
        & 18.37 & 420 \\

        \hline
8
        & 7.62& 198
        & 6.79 & 182
        & 6.8 & 187
        & 7.32 & 188
        & 3.8 & 184
        & 4.18 & 179 \\

        \hline
9
        & 1.91 & 190
        & 2.05 & 190
        & 2.04 & 190
        & 2.03 & 192
        & 2.11 & 190
        & 2.06 & 190 \\

        \hline
10
        & 0.28 & 39
        & 0.39 & 45
        & 0.36 & 43
        & 0.38 & 42
        & 0.4 & 43
        & 0.01 & 44 \\

        \hline
11
        & 17.83 & 144
        & 16.51 & 137
        & 16.26 & 144
        & 17.11 & 141
        & 17.47 & 139
        & 17.24 & 139 \\

        \hline
12
        & 0.79 & 112
        & 0.77 & 113
        & 0.82 & 118
        & 0.82 & 118
        & 1.13 & 119
        & 0.13 & 126 \\

        \hline
13
        & 4.19 & 221
        & 4.88 & 229
        & 4.99 & 233
        & 4.98 & 230
        & 5.23 & 233
        & 5.57 & 235 \\

        \hline
14
        & 9.49 & 354
        & 8.01 & 357
        & 8.14 & 359
        & 8.29 & 358
        & 8.28 & 364
        & 8.32 & 361 \\

        \hline
        Network  
        & 6.17 & 454 
        & 5.76 & 473 
        & 5.91 & 466 
        & 6.31 & 475
        & \textbf{5.09} & \textbf{493} 
        & 5.95 & 478\\

        \hline
    \end{tabular}
}
\end{adjustbox} 
\end{center}
\vspace{-8pt}
\caption{\small{Average waiting time $\overline{W}~(s)$ and vehicle throughput $Q~(v/500s)$ for 14 individual intersections and entire network when RV rate = 80\%. The configurations~$2U+12S$,~$4U+10S$,~$8U+6S$, and~$10U+4S$ achieve lower network~$\overline{W}$ and higher network~$Q$, while~$6U+8S$ gets higher~$\overline{W}$ and higher~$Q$ compared to the baseline~$0U+14S$. For the entire network, the highest improvement for~$\overline{W}$ and~$Q$ both occur at~$8U+6S$ configuration. For intersection 6, the highest improvement occur with an 98.43\% reduction in~$\overline{W}$. $8U+6S$ achieves the lowest~$\overline{W}$ of 5.09~$s$ across the entire network and has the lowest $\overline{W}$ at four of the 14 intersections, while the baseline and~$2U+12S$ rank second with three intersections at the same time. These results make~$8U+6S$ the best configuration for minimizing delays when RV rate = 80\%. For~$Q$,~$8U+6S$ also achieves the highest network-wide value of 493~$v/500s$ and has the highest $Q$ at two intersections, while the~$10U+4S$ ranks first with four intersections. These results indicate that~$8U+6S$ is the best configuration for maximizing network-wide throughput.}}
\label{tab:80results}
\vspace{-10pt}
\end{table*}

\begin{figure*}
    \centering
    \includegraphics[width=\linewidth]{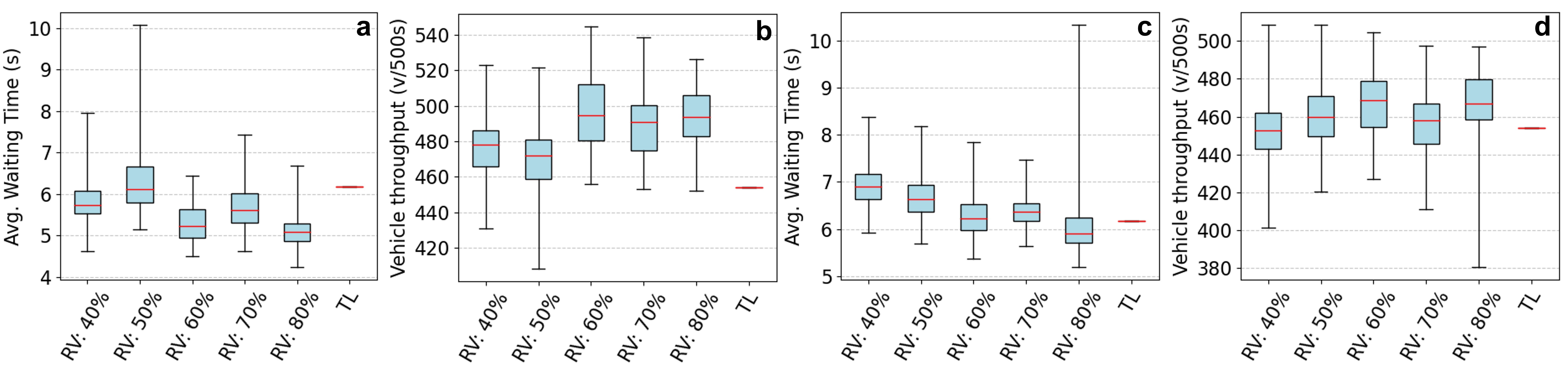}
    \caption{\small{Average waiting time~$\overline{W}~(s)$ and vehicle throughput~$Q~(v/500s)$ in the~$8U+6S$ configuration (a,b) and the~$4U+10S$ configuration (c, d) for the entire network. For the~$8U+6S$ configuration, our method consistently outperforms the baseline for all five RV penetration rates, in both~$\overline{W}$ and~$Q$. (a):~$\overline{W}$ decreases for 40\%, 60\% and 80\%, while 50\% and 70\% show higher values. (b):~$Q$ increases for 40\%, 60\% and 80\%, while 50\% and 70\% show lower values. For the~$4U+10S$ configuration, our method outperforms the baseline when RV rate = 80\% in~$\overline{W}$ and outperforms the baseline when RV rates are above 40\% in~$Q$. (c):~$\overline{W}$ decreases for 40\%, 50\%, 60\% and 80\%, while 70\% shows higher values. (d):~$Q$ increases for 40\%, 50\%, 60\% and 80\%, while 70\% shows lower values.}}
    \label{fig:Fig_2}
    \vspace{-10pt}
\end{figure*}

In Fig.~\ref{fig:Fig_2}, the average waiting time and vehicle throughput are analyzed for the~$8U+6S$ and~$4U+10S$ configurations across the entire network. In the~$8U+6S$ setup, our method consistently outperforms the baseline across all five RV penetration rates for both metrics. The waiting time decreases at 40\%, 60\%, and 80\%, while 50\% and 70\% show higher values. Similarly, throughput improves at 40\%, 60\%, and 80\%, but declines at 50\% and 70\%. In the~$4U+10S$ configuration, the method reduces waiting time significantly at an 80\% RV rate and improves throughput when the RV rate exceeds 40\%. Waiting time decreases at 40\%, 50\%, 60\%, and 80\%, but increases at 70\%. Throughput follows a similar trend, rising at 40\%, 50\%, 60\%, and 80\%, while declining at 70\%. For overall network performance across all five configurations, the average waiting time~$\overline{W}$ generally reduces as the RV penetration rate increases from 40\% to 80\%, indicating improved traffic efficiency. The vehicle throughput~$Q$ increases across all configurations, suggesting more vehicles successfully reach their destinations. The most significant improvement is observed in the~$8U+6S$ configuration when RV rate = 80\%, where \( \overline{W} \) is reduced from 6.17~$s$ to 5.09~$s$, and~$Q$ increases from 454~$v/500s$ to 493~$v/500s$, compared to the baseline ($0U+14S$). This suggests that controlling some intersections with RVs while keeping others managed by traffic signals improves overall traffic performance when the RV rate is high enough.

\begin{figure*}
    \centering
    \includegraphics[width=\linewidth]{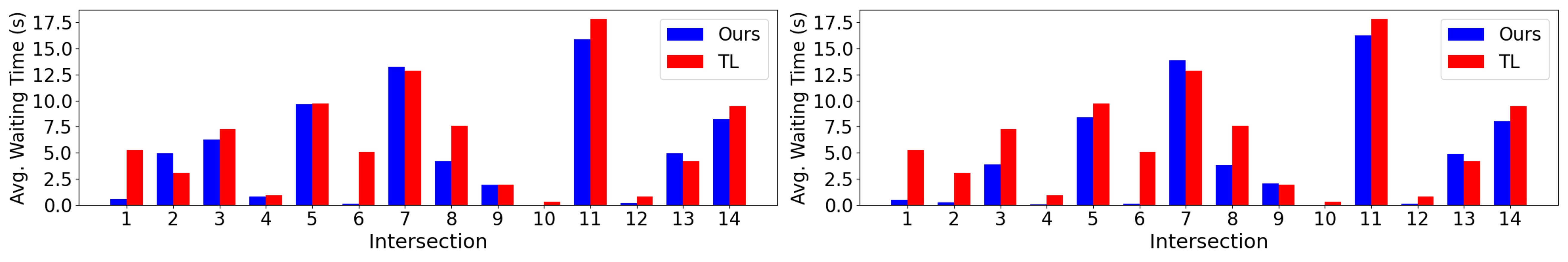}
    \vspace{-20pt}
    \caption{\small{Average waiting time~$\overline{W}~(s)$ for 14 intersections when RV rate = 50\% (upper figure) and 80\% (lower figure). Ours is the lowest value among five configurations from~$2U+12S$ to~$10U+4S$. At the 50\% RV rate, intersections 1, 4, 5, 6, 8, 10, 11, 12 and 14 show that the best~$\overline{W}$ is lower than the traffic light control (TL). At the 80\% RV rate, intersections 1, 2, 3, 4, 5, 6, 8, 10, 11, 12 and 14 show that the best~$\overline{W}$ is lower than TL. These results indicate that our method can reduce vehicle waiting times at most intersections compared to traffic light control. Additionally, the performance of our method tends to improve as the RV rate increases from 50\% to 80\%.}}
    \label{fig:Fig_3}
    \vspace{-16pt}
\end{figure*}

Fig.~\ref{fig:Fig_3} shows the average waiting time~$\overline{W}$ for 14 intersections when RV rate = 50\% and 80\%. At 50\% RV rate, nine intersections achieve lower~$\overline{W}$ than traffic light control, increasing to 11 intersections at 80\% RV rate. These results demonstrate that our method reduces vehicle waiting times at most intersections and improves as the RV rate rises. At each individual intersection, our method's performance varies, and the analysis of all 14 intersections reveals several trends. 
To start with, at intersections 1, 6, 8, 11, and 14, the average waiting time is consistently lower across all control configurations and RV penetration rates, with intersection 1 achieving waiting times up to 10 times lower than the baseline; however, in some cases the vehicle throughput, although increasing with higher RV rates, remains slightly below the baseline. At intersections 3 and 5, high RV penetration of 60\% and 80\% respectively ensures that waiting times are better than the baseline across all configurations, yet throughput, despite rising with RV rate, still falls marginally short of the baseline level. In contrast, intersections 2, 4, 7, 9, 10, 12, and 13 show mixed waiting time performance, with some control configurations outperforming the baseline and others not. Notably, intersection 10 only exhibits improved waiting time under the~$10U+4S$ configuration, although its throughput is consistently higher than the baseline across all scenarios. Furthermore, intersections 4, 9, 12, and 13 generally achieve higher throughput than the baseline; for instance, intersection 9 shows throughput that increases with RV rate and ultimately exceeds the baseline, even as waiting time improvements vary. Overall, while advanced control strategies tend to reduce waiting times at most intersections, the corresponding gains in vehicle throughput are less uniform, reflecting a complex interplay between local traffic dynamics and control configuration.

\subsection{Results At Various RV Rates}
Tables~\ref{tab:50results} and~\ref{tab:80results} show the results of traffic performance at the 14 intersections and the entire network under six network control configurations at RV rate 50\% and 80\%, respectively. 
In each table, rows 1--14 show individual intersection results, while the last row reports the entire network results. 

When the RV rate is at 50\%, the configurations~$2U+12S$ and~$8U+6S$ achieve lower network~$\overline{W}$ and higher network~$Q$, while~$4U+10S$,~$6U+8S$ and~$10U+4S$ get higher~$\overline{W}$ and higher~$Q$ compared to the baseline~$0U+14S$. Over the entire network, the highest improvement (bold values in the Table) for the two metrics occur at~$2U+12S$ and~$10U+4S$ configuration. Our method impacts individual intersections differently: some achieve lower~$\overline{W}$ and others experience higher~$\overline{W}$. 
For intersection 6, the highest improvement happens with an 98.04\% reduction in~$\overline{W}$. $2U+12S$ and~$8U+6S$ outperform the baseline with waiting times of 5.86~$s$ and 6.11~$s$ and throughput of 470~$v/500s$ and 471~$v/500s$, respectively, but the~$6U+8S$ and~$10U+4S$ configurations exhibit increased waiting times (7.92~$s$ and 7.02~$s$) despite moderate throughput gains. 

When the RV rate reaches 80\%, the~$8U+6S$ configuration achieves the best performance overall with an average waiting time of 5.09~$s$ and throughput of 493~$v/500s$, while~$2U+12S$,~$4U+10S$ and~$10U+4S$ also improve waiting times and throughput compared to the baseline. For the network, the highest improvements for the two metrics, both occur at~$8U+6S$ configuration. Our method still impacts individual intersections differently: some achieve lower~$\overline{W}$ and the others experience higher~$\overline{W}$. For intersection 6, the highest improvement occurs with an 98.43\% reduction in~$\overline{W}$. 


We also conduct similar experiments when RV rate = 40\%, 60\% and 70\%. At the 40\% RV rate, most configurations improve performance relative to the baseline of 6.17 and 454~$v/500s$, although the degree varies. The~$2U+12S$ configuration reduces waiting time to 5.89 $s$ and increases throughput to 474~$v/500s$, while the~$8U+6S$ setup further lowers waiting time to 5.74 $s$ and boosts throughput to 477~$v/500s$. The~$10U+4S$ configuration also shows strong performance with a waiting time of 5.94 $s$ and the highest throughput of 483~$v/500s$. In contrast, the~$4U+10S$ configuration underperforms in waiting time (6.9 $s$) with a throughput close to baseline (452~$v/500s$), and the~$6U+8S$ delivers moderate gains (6.43 $s$ and 468~$v/500s$). In general, these results suggest that at a 40\% RV penetration, configurations such as~$2U+12S$,~$8U+6S$, and~$10U+4S$ could reduce delays and improve throughput compared to the baseline, while the others show mixed performance. 

At the 60\% RV rate, improvements are more obvious with~$10U+4S$ and~$8U+6S$ reducing waiting times to 5.51~$s$ and 5.22~$s$ and achieving throughput of 481~$v/500s$ and 492~$v/500s$, while~$6U+8S$ and~$4U+10S$ remain near or slightly above baseline in waiting time. For intersection 6, there is an improvement with an 98.82\% reduction in~$\overline{W}$. For the 70\%, although~$2U+12S$ and~$8U+6S$ continue to offer lower waiting times (5.9~$s$ and 5.6~$s$) and improved throughput (471~$v/500s$ and 490~$v/500s$), the configurations~$4U+10S$,~$6U+8S$ and~$10U+4S$ get higher~$\overline{W}$ and higher~$Q$ compared to the baseline~$0U+14S$. The~$6U+8S$ configuration again shows degraded waiting times (7.85~$s$) and slightly lower throughput, and~$10U+4S$'s waiting time increases to 6.23~$s$. For intersection 6, there is an improvement with an 98.24\% reduction in~$\overline{W}$. The detailed results can be found at \href{https://github.com/cgchrfchscyrh/MixedTrafficControl_IROS/blob/main/Tables.pdf}{https://github.com/cgchrfchscyrh/MixedTrafficControl\_IR\allowbreak OS/blob/main/Tables.pdf}.

\section{Conclusion and Future Work} 
We propose the first application of decentralized multi-agent reinforcement learning for large-scale mixed-traffic intersection control, where RL-driven RVs regulate some intersections while traffic signals control others. Unlike previous studies focused solely on signalized or unsignalized control, we explore their coexistence to optimize urban mobility. Our MARL-based approach enables RVs to adapt dynamically to local traffic conditions, offering a flexible alternative to static signals. Experiments on a 14-intersection real-world network (Colorado Springs, CO, USA) compare five configurations of mixed control against a fully signalized baseline. Traffic performance is evaluated using average waiting time and vehicle throughput. The results show that at 80\% RV penetration, $\overline{W}$ decreases by 20\% (from 6.17~$s$ to 5.09~$s$) while $Q$ increases by 9\% (from 454 to 493 vehicles per 500 seconds). This research examines the transition from conventional traffic signals to autonomous intersection management. With the rise of self-driving technology, our results suggest that integrating autonomous control can enhance traffic flow while reducing the need for static signals. By demonstrating these benefits, our study lays the groundwork for future developments in signal-free intersection systems.

While our method shows promise in managing large-scale mixed traffic with combined control strategies, several limitations remain. First, although we adopt a Rainbow DQN-based MARL framework, it represents a relatively standard approach. In future work, we aim to explore more advanced architectures, such as graph-based MARL, attention mechanisms for agent coordination to improve learning efficiency and coordination under sparse and delayed rewards. Second, in practice, decentralized learning may strain onboard computational resources, and vehicle communication is often unreliable due to latency and packet loss. We plan to incorporate realistic V2X models and investigate model compression for efficient policy deployment. Third, the poor performance of the~$6U+8S$ configuration suggests that network geometry, vehicle route distribution, and suboptimal intersection selection may hinder congestion mitigation. To address this, we plan to explore adaptive intersection selection. 
We also plan to investigate how different origin-destination demands affect RV-controlled intersections. Finally, our study focuses solely on intersections, overlooking other road structures like one-way streets and roundabouts. Expanding our analysis to diverse network layouts will enhance the adaptability of our method to real-world traffic management.

\section*{Acknowledgment}

This research is funded by the National Science Foundation (NSF) via Grant 2524239, 2129003, and 2222810. The authors gratefully acknowledge NSF’s support. The authors would also like to thank NVIDIA and the Tickle College of Engineering at University of Tennessee, Knoxville for their support. 








\bibliographystyle{IEEEtran}
\bibliography{export}
\end{CJK}
\end{document}